\documentclass[10pt,twocolumn,letterpaper]{article}

\usepackage{wacv}              

\usepackage{array}
\usepackage{colortbl}
\definecolor{hourBlue}{HTML}{9BBDE0}
\definecolor{hourBlueLight}{HTML}{CBDEF0}
\definecolor{hourPeach}{HTML}{F0BAA0}
\definecolor{hourMint}{HTML}{B5D8C2}
\definecolor{hourLavender}{HTML}{BEAFD8}
\definecolor{hourNeutral}{HTML}{B6B8B1}
\definecolor{hourUnknown}{HTML}{DDDCD6}
\definecolor{hourBlueInk}{HTML}{52759A}
\definecolor{hourGreenInk}{HTML}{547C65}
\definecolor{hourCoralInk}{HTML}{A56152}
\definecolor{hourInk}{HTML}{333941}
\definecolor{hourHeader}{HTML}{EDE8F5}
\definecolor{hourOursBg}{HTML}{EAF1F9}
\definecolor{hourPerfectBg}{HTML}{F2EEF8}
\definecolor{hourBaselineBg}{HTML}{F3F3F0}
\definecolor{hourGroupBg}{HTML}{EDF5EF}
\definecolor{hourPeachBg}{HTML}{FBE8DB}
\definecolor{hourMintBg}{HTML}{E6F2E9}
\definecolor{hourBlueBg}{HTML}{E5EFF8}
\definecolor{hourLavenderBg}{HTML}{EAE4F5}
\definecolor{hourCoralBg}{HTML}{F9E2DB}
\colorlet{wacvblue}{hourBlueInk}

\newcommand{\revcolor}{\color{blue}}                  
\newcommand{\rev}[1]{{\revcolor#1}}
\newcommand{\supp}[1]{Supp.~#1}                        

\newcommand{\ci}[2]{\,{\scriptsize[#1,\,#2]}}          
\newcommand{\fix}[1]{Fix~#1}                           
\newcommand{\fixes}[1]{Fixes~#1}                       
\newcommand{\para}[1]{\noindent\textbf{#1}\;}

\ifdefined\XeTeXrevision
  \usepackage{fontspec}
\fi

\usepackage[pagebackref,breaklinks,colorlinks,allcolors=wacvblue]{hyperref}

\def\wacvPaperID{11} 
\def\confName{WACV}
\def\confYear{2027}

\title{How Well Do Vision-Language Scene Graphs Serve\\as Memory for Embodied Question Answering?}

\author{Divyansh Pandey\textsuperscript{*}\quad
Shubh Garg\textsuperscript{1,*}\quad
Vedant Khandelwal\textsuperscript{2}\quad
Niyati Rawal\textsuperscript{1,3}\quad
Amit Sheth\textsuperscript{1}\\[0.3em]
{\small\textsuperscript{1}Indian AI Research Organisation (IAIRO), GIFT City, Gujarat\quad
\textsuperscript{2}AI Institute, UofSC (AIISC), 1112 Greene St., Columbia}\\
{\small\textsuperscript{3}BITS Pilani\quad
\textsuperscript{*}Equal contribution}\\[0.2em]
{\footnotesize\texttt{\{divyanshpandey0108@gmail.com, divyansh@upailabs.org\}, \{shubh.garg@iairo.ai, shubhgarg265@gmail.com\},}}\\
{\footnotesize\texttt{kh.vedant02@gmail.com, niyati.rawal@iairo.ai, \{amit.sheth@iairo.ai, amitpsheth@gmail.com\}}}}

\begin{document}
\twocolumn[{%
\renewcommand\twocolumn[1][]{#1}%
\maketitle
\vspace{-2.8em}
\begin{center}
\includegraphics[width=0.93\textwidth]{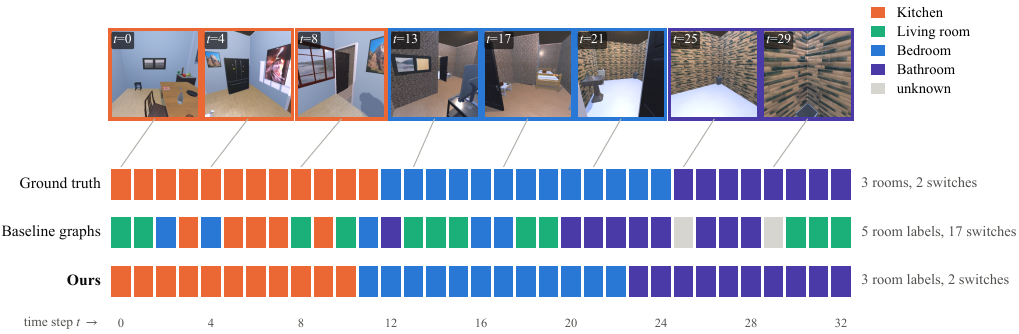}
\captionof{figure}{\textbf{One episode, three memories.} The agent crosses kitchen\,$\rightarrow$\,bedroom\,$\rightarrow$\,bathroom in 33 steps (top: eight frames; colour: true room). Baseline graphs give five labels and 17 switches; segments from simulator door crossings recover the three-room route, with boundaries one or two steps early (test episode \texttt{ep\_0003}).}
\label{fig:short-teaser}
\end{center}%
}]

\begin{abstract}
Embodied QA needs memory, but answer accuracy mixes retained evidence, reading errors and question priors. For a simulated multi-room scene-graph pipeline we define \emph{answerability}: the fraction of questions a fixed rule-based reader answers correctly from the graph alone. The baseline recovers 24.1\% of visible object types and its room labels flicker. Four targeted repairs double recall, raise answerability from 27.2\% to 37.4\%, and improve retrained QA from 71.4\% to 75.9\%, yet better object F1 can lower answerability. A question-type prior beats this reader on every predicted graph. In one real-video test, graph memory lowers accuracy. These exploratory findings favour scoring retained evidence alongside accuracy.
\end{abstract}

\begin{figure*}[t]
\centering
\includegraphics[width=\textwidth]{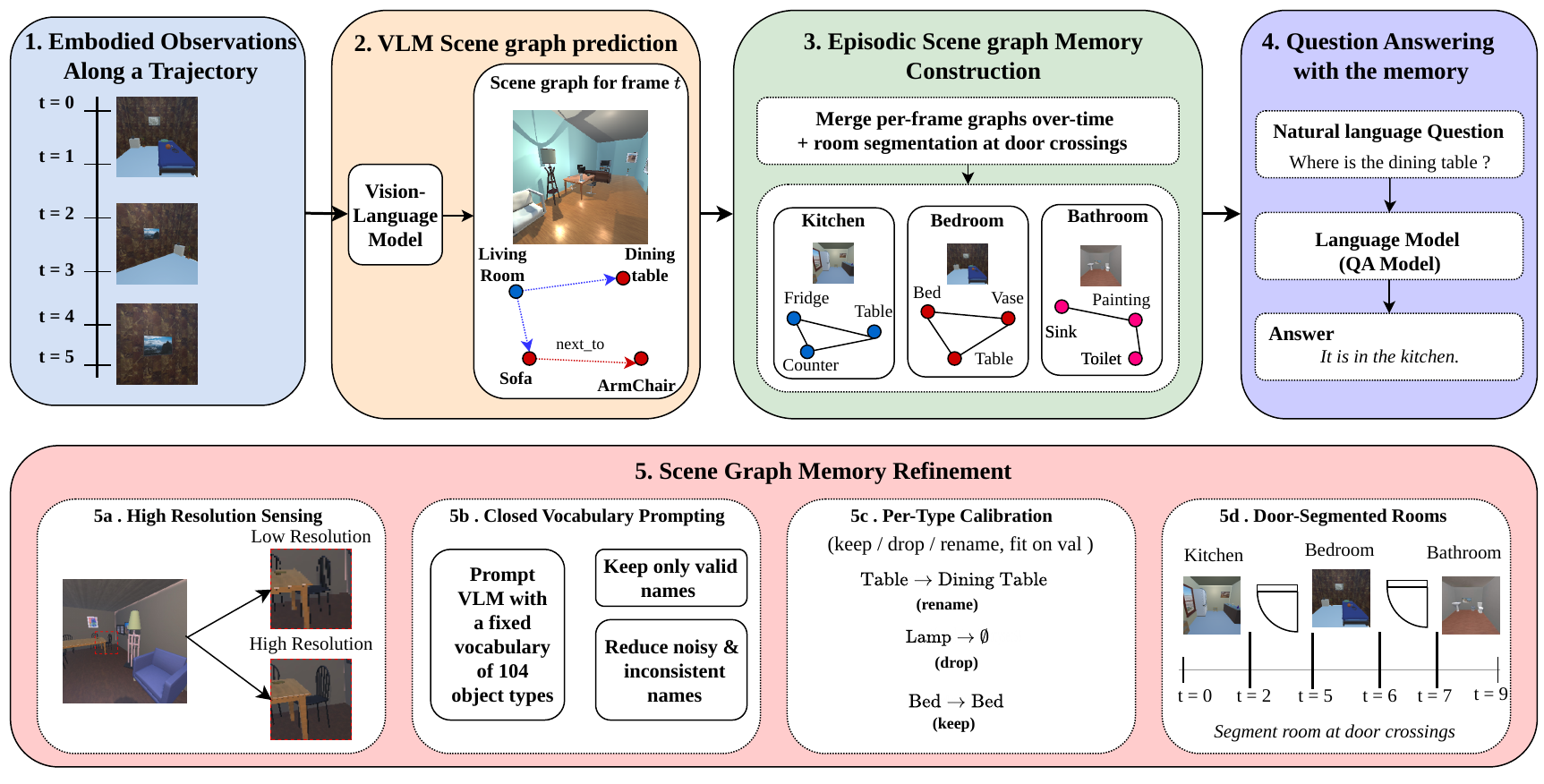}
\caption{\textbf{Memory architecture and four repairs.} Per-frame graphs are merged into episodic memory for QA. Fixes 1 and 2 change resolution and prompting; Fix 3 calibrates object types; Fix 4 segments rooms at door crossings (shown). The prompt adds Shelf and ShelvingUnit to the 104 types shown. The QA example uses a blank answer-time image.}
\label{fig:short-pipeline}
\end{figure*}

\section{Introduction}
\label{sec:intro}

An embodied assistant answering questions about a home must remember what it saw after leaving a room~\cite{das2018eqa,majumdar2024openeqa,yadav2026findingdory}. A common approach predicts each frame's objects, relations and room as a scene graph, then lets a language model read the merged graph~\cite{saxena2025grapheqa,gu2024conceptgraphs,werby2024hovsg}. Final-answer accuracy mixes memory contents, the reader's ability to use them, and question priors~\cite{thomason2019shifting,goyal2017vqa}. A correct guess can conceal missing evidence; a reading error can conceal useful memory.

We audit a multi-room QA pipeline built on MomaGraph-R1~\cite{ju2026momagraph}. Its benchmark, baseline graphs, training recipe and no-memory/baseline QA runs are inherited (\supp{S1, S7, S11}); our contribution is a diagnostic evaluation of memory evidence rather than a new memory method. We define \emph{answerability} using a fixed rule-based reader, with abstentions counted as wrong. Four targeted repairs act as diagnostic interventions: higher object F1 can reduce question recovery, and better room continuity can hurt object--room queries (\cref{fig:short-teaser,fig:short-pipeline}).

\par
\para{Related work.}
Across interactive/photorealistic QA~\cite{gordon2018iqa,wijmans2019eqa}, long-horizon memory~\cite{zhai2025memoryeqa,hu2025mem,anwar2025remembr}, egocentric video~\cite{grauman2022ego4d} and situated 3D QA~\cite{azuma2022scanqa,ma2022sqa3d}, answer accuracy evaluates end performance. Question-prior controls~\cite{agrawal2018vqacp,brown2025benchmark} motivate separating that score from retained visual evidence. Saha \etal~\cite{saha2026newevidence} likewise score evidence selection apart from final answers; we score evidence retention.

Scene graphs encode building hierarchies~\cite{armeni2019sg,rosinol2020dsg,hughes2022hydra}, reconstructed objects~\cite{wald2020learning} and open-vocabulary semantics~\cite{koch2024open3dsg,maggio2024clio} for planning~\cite{rana2023sayplan}, navigation~\cite{yin2024sgnav} and QA~\cite{ali2025graphpad,saxena2025grapheqa}. Image memories retain another form of evidence~\cite{yang2025mem,ren2024explore}; we examine what graph evidence supports before a trained task reader.

Image scene-graph generation~\cite{xu2017imp,zellers2018motifs,tang2020unbiased} and panoptic graphs~\cite{yang2022psg} target object/relation recovery using annotations such as Visual Genome~\cite{krishna2017vg}. Our fixed reader tests whether recovered content answers questions, connecting unanswerable-QA analysis~\cite{rajpurkar2018know} with scene-graph evaluation gaps~\cite{rotondi2026survey}. \supp{S14} develops these distinctions.

\begin{table}[t]
\centering
\caption{\textbf{Fixed-reader rules.} Missing evidence or an ambiguous decision yields no answer. Full question examples are in \supp{S11}.}
\label{tab:short-rules}
\footnotesize
\setlength{\tabcolsep}{3pt}
\begin{tabular}{@{}l>{\raggedright\arraybackslash}p{5.30cm}@{}}
\toprule
Question type & Rule applied to the graph\\
\midrule
Order & Compare first sightings of the two types.\\
Presence-change & Yes if the last sighting precedes the final memory frame.\\
Rooms-visited & Count distinct room labels.\\
Persistence & Count rooms containing the queried type.\\
Room ID & Return the type's room if unique.\\
Existence & Pick the unique option present in the asked room.\\
Counting & Use the largest count in one frame in that room.\\
Global count & Sum per-type maximum counts in the category.\\
\bottomrule
\end{tabular}
\end{table}

\begin{table*}[t]
\centering
\caption{\textbf{Repair ladder} (200 test episodes; 8{,}135 memory frames). P/R: object-type precision/recall (\%); answerability includes 95\% CIs. Ours uses all four fixes with full calibration; the conservative variant is not QA-retrained. The vocabulary map normalises baseline names without VLM inference. Additional episode metrics and paired step differences are in \supp{S13}.}
\label{tab:short-ladder}
\footnotesize
\setlength{\tabcolsep}{6pt}
\begin{tabular}{@{}lccccc@{}}
\toprule
Memory graphs & Object P & Object R & Object F1 & Room accuracy & Answerability\\
\midrule
Baseline (256\,px, open prompt) & \cellcolor{hourBlue!25.19}45.8 & \cellcolor{hourBlue!13.26}24.1 & \cellcolor{hourBlue!17.38}31.6 & \cellcolor{hourBlue!24.86}45.2 & \cellcolor{hourBlue!14.96}27.2\ci{24.1}{30.4}\\
\quad + vocabulary map & \cellcolor{hourBlue!31.18}56.7 & \cellcolor{hourBlue!15.56}28.3 & \cellcolor{hourBlue!20.79}37.8 & \cellcolor{hourBlue!24.86}45.2 & \cellcolor{hourBlue!15.95}29.0\ci{25.6}{32.5}\\
\fixes{1+2}: 512\,px, closed prompt & \cellcolor{hourBlue!26.12}47.5 & \cellcolor{hourBlue!27.55}50.1 & \cellcolor{hourBlue!26.84}48.8 & \cellcolor{hourBlue!27.00}49.1 & \cellcolor{hourBlue!19.91}36.2\ci{33.1}{39.4}\\
\quad + \fix{3}: calibration & \cellcolor{hourBlue!36.13}65.7 & \cellcolor{hourBlue!27.06}49.2 & \cellcolor{hourBlue!30.96}56.3 & \cellcolor{hourBlue!27.00}49.1 & \cellcolor{hourBlue!19.36}35.2\ci{31.9}{38.6}\\
\quad + \fix{4}: door segments (\textbf{ours}) & \cellcolor{hourBlue!36.13}65.7 & \cellcolor{hourBlue!27.06}49.2 & \cellcolor{hourBlue!30.96}56.3 & \cellcolor{hourBlue!39.49}71.8 & \cellcolor{hourBlue!20.57}37.4\ci{34.2}{40.6}\\
Ours, conservative calibration & \cellcolor{hourBlue!31.41}57.1 & \cellcolor{hourBlue!29.54}53.7 & \cellcolor{hourBlue!30.47}55.4 & \cellcolor{hourBlue!39.49}71.8 & \cellcolor{hourBlue!21.95}39.9\ci{36.5}{43.4}\\
\bottomrule
\end{tabular}
\end{table*}

\section{Setting and protocol}
\label{sec:setting}

\para{Benchmark and dataset.}
In each episode an agent in a ProcTHOR-10k house~\cite{deitke2022procthor} (AI2-THOR~\cite{kolve2017ai2thor}) walks from its spawn point to a target object in another room, through two to seven rooms of four types (Bathroom, Bedroom, Kitchen, LivingRoom). Each step stores a 256\,px RGB frame, the agent's room and the simulator's visible-object record (type, instance ID and room of each visible object); episodes average 41.7 frames (10--116). Episodes are balanced over one, two, three and four or more door crossings. Templates write four multiple-choice questions per episode (two to four options; chance 33\%) from the records of six sampled frames, the \emph{question strip}. Eight types cover sequence queries (Order, Presence-change, Rooms-visited; 291 test questions) and set queries about objects and rooms (509; \cref{tab:short-rules}). Splits are house-disjoint: 5{,}000/500/200 episodes from 1{,}359/151/53 houses, with 800 questions and 8{,}335 frames on test. \supp{S11} gives dataset statistics and examples. Memory contains all frames except the last question-time view: 8{,}135 test frames.

\para{Memory and QA.}
MomaGraph-R1, based on Qwen2.5-VL-7B~\cite{bai2025qwen25vl}, predicts objects, relations and rooms (\cref{fig:short-pipeline}). A fixed renderer merges graphs into room-grouped text with visit order, first/last sightings, maximum simultaneous counts and relations. It drops walls, doors and floors, retaining at most 12 object types per room by sighting frequency and 24 relations per frame (\supp{S1}). Graphs represent types, not tracked instances. MomaGraph-R1 is QA-trained with LoRA~\cite{hu2022lora} for 6{,}000 optimiser steps, effective batch 16. Its answer-time image is blank; it reads the question, options and memory text. Training without memory supplies the question-only baseline. Runs share an auxiliary next-action objective and final-checkpoint evaluation (\supp{S7}).

\para{Protocol.}
Simulator truth scores graphs and fits calibration/room naming on training/validation data. Only door segmentation additionally uses exact crossings; \emph{perfect} rows use simulator graphs. A 300-test-frame prompt/resolution pilot informed development; test scores were visible during the conservative-calibration tie-break (\supp{S10}). We therefore treat all test comparisons as exploratory. 95\% confidence intervals (CIs) capture episode variability but not this selection effect: 20{,}000 paired bootstrap resamples of whole test episodes~\cite{efron1993bootstrap}, averaging two QA seeds first.

\section{Measuring retained evidence}
\label{sec:measuring}

\para{Answerability.}
A fixed reader applies \cref{tab:short-rules} to the episodic graph and returns an option or abstains. Answerability is the fraction answered correctly; abstentions count as wrong. The rules have no learned prior, free parameters or test tuning. This scores graph recovery under specified rules; QA evaluates a trained reader of the rendered text. Object-type precision/recall/F1 compare predicted and visible types after benchmark-vocabulary name matching; room accuracy measures label correctness.

\para{Reference graphs.}
We separate room filing from temporal coverage using simulator-perfect graphs. \emph{Question strip, physical rooms}: 97.5\% (CI [96.4, 98.5]), with 100\% on seven types and 82.8\% on Rooms-visited. \emph{Memory frames, physical rooms}: 80.4\% [77.8, 83.0]. \emph{Memory frames, agent's room}: 59.9\% [56.8, 63.1]. The 20.5-point filing cost arises when doorway sightings are assigned to the agent's room rather than the object's physical room. Extra memory frames can change counts and sighting times; room-type labels can merge distinct physical rooms. These constructions characterize the benchmark and reader, rather than universal memory bounds.

\section{Diagnosis and targeted repairs}
\label{sec:failures}

Baseline graphs miss objects and lose room continuity. Over all 8{,}335 test frames (including the last frames excluded from \cref{tab:short-ladder}), object-type recall is 24.1\%; small-object recall is only 8.3\% even within 1.5\,m. Of false positives, 57\% are types absent from the entire episode. Room labels are correct in 45.5\% of frames and switch 13.6 times per episode, versus 2.1 true label changes. Relation recall is 1.1\%, and graphs lack the instance identities counting needs. Visual examples and the full breakdown are in \supp{S12}.

Naming and hallucination errors recur systematically, consistent with object-hallucination studies in captioning and VLMs~\cite{rohrbach2018hallucination,li2023pope}. The model often calls a dining table a Table or SideTable; six furniture types account for 61\% of false positives. Requiring repeated sightings does not remove these persistent errors. These patterns motivate prompt redesign and per-type calibration, while room flicker motivates segment-level room decisions.

\par
\para{\fixes{1+2}: resolution and prompting.}
To recover small-object detail~\cite{wu2024vstar} and control naming, we re-render stored poses at 512\,px and prompt with 106 types: the benchmark's 104 plus Shelf and ShelvingUnit. Baseline graphs use 256\,px frames and a prompt conditioned on the episode's first question (graphs shared by its four questions) that lists only navigation structures outside the target room. The replacement drops both and requests all visible non-architectural objects, so it tests the prompt bundle, not vocabulary alone. In the 300-frame pilot, the prompt adds 16.0 recall points and higher resolution 12.3 more (\supp{Tab.~S8}). A post-hoc Grounding DINO~\cite{liu2024groundingdino} pilot scored below the VLM (\supp{S6}).

\para{\fix{3}: per-type calibration.}
We greedily fit keep/drop/rename rules to validation $F_{0.5}$, which weights precision above recall. The 28 rules rename Desk, Table and CoffeeTable to DiningTable and drop 25 types. A \emph{conservative} variant uses validation answerability with a precision tie-break (\supp{S10}); it retains the renames and drops only nine types with precision below 10\%. Rules calibrate type decisions rather than confidence scores~\cite{guo2017calibration}.

\para{\fix{4}: door-segmented rooms.}
Following room-based geometric mapping~\cite{bormann2016room,hughes2022hydra}, consecutive agent door crossings delimit segments. Logistic regression names each segment from calibrated objects and VLM room votes, with training-data fitting and validation-selected regularisation. Repeated room types receive new numbers; these identify segments rather than physical-room revisits. A real agent would have to estimate these privileged crossings. \Cref{fig:short-pipeline} locates the interventions; details are in \supp{S1--S3}.

\section{Results}
\label{sec:results}
\begin{table}[t]
\centering
\caption{\textbf{Retrained QA} (800 questions, two-seed mean). $\Delta$: paired change from baseline with 95\% CI. Perfect graphs use simulator records.}
\label{tab:short-qa}
\footnotesize
\setlength{\tabcolsep}{4pt}
\begin{tabular}{@{}lcc@{}}
\toprule
Training/test memory & Accuracy (\%) & $\Delta$ (points)\\
\midrule
None & \cellcolor{hourBlue!35.64}64.8 & \cellcolor{hourPeach!24.27}$-$6.62\ci{$-$9.31}{$-$3.94}\\
Baseline graphs & \cellcolor{hourBlue!39.27}71.4 & ref.\\
Ours & \cellcolor{hourBlue!41.74}75.9 & \cellcolor{hourBlue!16.28}\textbf{+4.44}\ci{+2.19}{+6.69}\\
Perfect graph & \cellcolor{hourBlue!45.87}83.4 & \cellcolor{hourBlue!44.00}\textbf{+12.00}\ci{+9.06}{+14.88}\\
\bottomrule
\end{tabular}
\end{table}

\para{Graph quality and usefulness.}
The repairs double recall and raise answerability by +10.12\ci{+6.25}{+14.00} points (\cref{tab:short-ladder}); conservative calibration gains +12.62\ci{+8.38}{+16.88}. Most of the gain precedes door segmentation: \fixes{1--3}, without door segmentation, add +8.00\ci{+4.25}{+11.75}, or +10.50\ci{+6.62}{+14.25} conservatively. Door segmentation gets the full-episode room count right in 199/200 episodes, versus 46/200 for baseline, but its overall answerability increment, +2.12\ci{$-$0.88}{+5.25}, includes zero.

\para{Precision can conflict with usefulness.}
Calibration raises object F1 from 48.8 to 56.3 yet lowers answerability from 36.2 to 35.2. It removes types needed by Order questions, reducing that type's score from 46.6 to 32.9; conservative calibration restores 42.5. Door segmentation also trades off tasks: Rooms-visited improves from 32.8 to 59.5, but Existence falls from 40.7 to 24.8 because agent-room filing creates ambiguous object--room assignments. Other types are in \supp{Tab.~S3}.

\para{Sensitivity to door events.}
Over 10 corruption seeds, timing offsets up to eight frames cost only 0.45 points. Missing or adding 10\% of crossings lowers answerability to 36.1 or 35.6, respectively. At 25\% errors, both fall below the image-only configuration (\supp{S5}).

\para{A memory-free prior.}
The most frequent training answer value per question type scores 49.4\%, exceeding every predicted graph read alone. Our reader abstains on 44.6\% of questions; using the prior only on abstentions raises its score to 58.9\%, versus 53.4\% for baseline graphs with the same fallback. The paired gain remains +5.50\ci{+2.12}{+8.88}. \mbox{On Counting, always answering 2} scores 63.1\%, whereas predicted graphs score only 5--11\%. Missing objects and instance identities cost the reader most here.

\para{Does better memory help QA?}
Retraining on repaired graphs improves accuracy by +4.44 points (\cref{tab:short-qa}), concentrated in sequence questions (+11.17\ci{+6.68}{+15.71}); set questions change by +0.59\ci{$-$1.80}{+2.96}. \rev{Test-time swaps give smaller gains, and perturbing memory lowers accuracy} (\supp{S9.3, S13}).

\begin{table}[!ht]
\centering
\caption{\textbf{Real-video memory} (VSI-Bench debiased, Qwen3-VL-8B). Overall scores include paired scene-bootstrap 95\% CIs.}
\label{tab:short-vsi}
\footnotesize
\setlength{\tabcolsep}{3pt}
\begin{tabular}{@{}lcc@{}}
\toprule
Input & Overall & Counting\\
\midrule
Frames only & \cellcolor{hourBlue!26.02}\textbf{47.3}\ci{44.9}{49.7} & \cellcolor{hourBlue!25.57}\textbf{46.5}\\
Frames + graph memory & \cellcolor{hourBlue!24.20}44.0\ci{41.5}{46.6} & \cellcolor{hourBlue!19.41}35.3\\
Frames + other scene's memory & \cellcolor{hourBlue!24.09}43.8\ci{41.5}{46.2} & \cellcolor{hourBlue!20.52}37.3\\
Memory only & \cellcolor{hourBlue!15.73}28.6\ci{26.7}{30.5} & \cellcolor{hourBlue!6.66}12.1\\
\bottomrule
\end{tabular}
\end{table}

\para{Real-video check.}
On debiased VSI-Bench~\cite{yang2025thinking,brown2025benchmark} (2{,}362 questions, 274 ScanNet, ScanNet++ and ARKitScenes scans~\cite{dai2017scannet,yeshwanth2023scannetpp,baruch2021arkitscenes}), MomaGraph-R1 extracts graphs from 16 frames per scan at 640\,px using the closed prompt and 65 question-vocabulary types. Qwen3-VL-8B~\cite{bai2025qwen3vl} answers with official prompts/scoring, given identical frames in every frame-based condition. Adding memory reduces the score from 47.3 to 44.0, a paired scene-bootstrap change of $-$3.32\ci{$-$4.81}{$-$1.80} (\cref{tab:short-vsi}). Counting falls from 46.5 to 35.3. Another scene's memory scores 43.8, statistically indistinguishable from correct memory, so the loss may come from adding memory text rather than its content. Longer input, conflicting entries or unused graph content could each explain this (\supp{S8, S9.4}). It uses one extractor and one answerer and omits \fixes{3+4}, so it does not test the full pipeline.

\section{Scope and conclusion}
\label{sec:limits}
\label{sec:conclusion}

The reader is hand-written for eight templated question types, so free-form questions are untested; door segmentation uses simulator crossings; and since development saw test data, the gains need a pipeline frozen on training/validation data and scored once on unseen houses. Counting and room-assignment errors motivate instance identities and physical object locations in 3D. The real-video result motivates testing richer graph extractors and estimated room boundaries.
Within these limits, better object F1 can still reduce question recovery: evaluating retained evidence reveals task trade-offs that recognition scores and answer accuracy alone conceal.


\clearpage
{
    \small
    \bibliographystyle{ieeenat_fullname}
    \bibliography{main}
}

\setcounter{section}{0}\setcounter{table}{0}\setcounter{figure}{0}
\renewcommand{\thesection}{S\arabic{section}}
\renewcommand{\thetable}{S\arabic{table}}
\renewcommand{\thefigure}{S\arabic{figure}}
\renewcommand{\theHsection}{S\arabic{section}}
\renewcommand{\theHtable}{S\arabic{table}}
\renewcommand{\theHfigure}{S\arabic{figure}}
\maketitlesupplementary
This supplement contains implementation details, additional diagnostics and analyses that extend the four-page paper. The main paper defines the reader, repairs and evaluation protocol and reports the central graph, QA and real-video results. Here S1--S3 give prompt excerpts and fitting details; S4--S6 add reader, corruption and detector diagnostics; S7--S10 document training and development; S11 adds benchmark and dataset statistics and examples; and S12--S14 give visual analysis, additional episode metrics and extended related work. Unless specified otherwise, intervals use the main paper's paired episode-bootstrap protocol.

\section{Prompts and memory text}
\label{sec:s-prompts}

\para{Baseline scene-graph prompt.}
The baseline prompt is reproduced below, with long lines wrapped and dashes written as ``--''; braces mark a slot that is filled in. The baseline graphs use this prompt, which follows MomaGraph-R1's task-oriented format~\cite{ju2026momagraph}. Its task instruction is the episode's first question, given as context only. 

{\scriptsize\begin{verbatim}
You are an embodied robot perception system. Construct a scene
graph of ONLY the CURRENT image from visible evidence.
Building the scene graph is the task -- do NOT answer the
question.
Task instruction (context only -- do NOT answer it):
{task_instruction}
Reason first inside
<SCENE_GRAPH_ANALYSIS>...</SCENE_GRAPH_ANALYSIS>, then emit
exactly one JSON object inside
<SCENE_GRAPH_JSON>...</SCENE_GRAPH_JSON>.
STAGE 1 -- ROOMS (judge from the IMAGE, never from the
instruction text):
- scene_type: the room you are CURRENTLY IN, inferred from the
  visible objects/layout. Exactly one of: "Bathroom",
  "Bedroom", "Kitchen", "LivingRoom".
- target_scene: the room implied by the task instruction
  (where the target object lives). Same 4 values, or
  "unknown". This is the ONLY place the room named in the
  instruction may go -- it must NEVER set scene_type.
STAGE 2 -- OBJECTS (nodes): every object ACTUALLY VISIBLE now
(furniture, appliances, fixtures, doors, windows, decorations,
and navigation structures like door/hallway/entrance). Number
duplicates (chair1, chair2). Do NOT hallucinate hidden
objects. If scene_type != target_scene, the target object is
NOT in this room -- do NOT invent it; list only the visible
navigation structures (door/hallway/entrance) that could lead
out.
STAGE 3 -- EDGES: relationships between visible objects only,
using exactly these values:
- spatial_relationships: left_of, right_of, in_front_of,
  behind, higher_than, lower_than, close, far, touching
- functional_relationship: openorclose, adjust, control,
  providepower, activate, pairwith, support
- action_type: navigate, press, rotate, pull, push, open,
  close, insert, pick, release
- function_type: navigation, open_close_control,
  device_control, power_supply, water_flow_control,
  parameter_adjustment, special_function, assembly
Output EXACTLY (valid JSON only; do NOT answer the question):
<SCENE_GRAPH_ANALYSIS>
One short paragraph: which room you are in and the visible
objects that prove it, whether the target room differs (so
navigation is required), and the key visible
objects/relationships.
</SCENE_GRAPH_ANALYSIS>
<SCENE_GRAPH_JSON>
{"scene_type": "...", "target_scene": "...", "nodes": [...],
"edges": [{"object1": "...", "object2": "...",
"spatial_relationships": [...], "functional_relationship":
"...", "action_type": "...", "function_type": "...",
"is_touching": true}]}
</SCENE_GRAPH_JSON>
\end{verbatim}}

\para{Closed-vocabulary prompt (\fix{2}).}
The object list and output schema are abbreviated in this excerpt:

{\scriptsize\begin{verbatim}
You are an embodied robot perception system. List the objects
visible in this image.
STAGE 1 - ROOM: scene_type is the room you are CURRENTLY IN,
judged only from the image. Exactly one of: "Bathroom",
"Bedroom", "Kitchen", "LivingRoom".
STAGE 2 - OBJECTS: list EVERY object you can see, including
SMALL objects (cups, books, bottles, boxes, pillows, remote
controls, food, drawers, shelves) and objects in the background
or through doorways. Use ONLY these exact type names:
AlarmClock, Apple, ArmChair, ... [all 106 types] ... WineBottle
If an object matches none of these names, leave it out. Do NOT
list walls, floors, doors, windows or hallways. Number duplicates
(Chair1, Chair2). Do not invent hidden objects.
STAGE 3 - EDGES: spatial relationships between listed objects,
using exactly: left_of, right_of, in_front_of, behind,
higher_than, lower_than, close, far, touching.
[output format: <SCENE_GRAPH_ANALYSIS> one short paragraph,
then <SCENE_GRAPH_JSON> with scene_type, nodes, edges]
\end{verbatim}}

The ellipsis abbreviates the complete type list; it is not part of the extraction prompt.

\para{Answer prompt.}
Every QA model answers from this prompt, with a blank image (\cref{sec:s-train}). The options are joined into one line (``A: 2 B: 3 C: 4''), and the model's answer is the highest-scoring letter among the options the question offers.

{\scriptsize\begin{verbatim}
### Task: Navigation has finished.
Answer the question using the memory collected during
navigation.
### Question: {question}
### Choices: {choices}
### Memory:
{memory text}
### Guidance:
Select the correct answer from the given options.
### Answer:
\end{verbatim}}

The no-memory model drops the two memory lines, and its second line reads ``Answer the question using the visual information collected during navigation.'' 

\para{Navigation prompt.}
Training also includes an auxiliary next-action loss (\cref{sec:s-train}). Its prompt pairs the current frame with that frame's scene graph, written as three lines (ROOM, up to eight OBJECTS under their raw names, RELATIONS), with doors and walls kept:

{\scriptsize\begin{verbatim}
### Task: Predict the next navigation action.
### Question: {question}
### Choices: {choices}
### Candidate Actions: (0) MoveAhead (1) RotateLeft
  (2) RotateRight (3) LookUp (4) LookDown (5) Done
### Predicted Scene Graph:
{current frame's scene graph}
### Guidance:
Use the scene graph and the question to choose the best next
action.
Choose Done only when sufficient information has been
gathered to answer the question.
### Action Token: <CLS>
\end{verbatim}}

\para{Memory text.}
The memory text has this form:

{\scriptsize\begin{verbatim}
ROOMS: {rooms in order of first visit}
VISITS (frames f0-f{N}): {room}@f{a}-f{b} -> {room}@f{c}-f{d}
  -> ...
{room}: {Type} x{n}[f{first}-f{last}],
  {Type}[f{first}-f{last}], ...
RELATIONS: {Type}-{relation}-{Type}; ...
\end{verbatim}}

Frame indices start at the episode's beginning. Multiplicity is printed only when above one. Functional tags, when available, follow the frame interval, and repeated relations are listed once. Free-text names are mapped to the benchmark vocabulary before serialization.

\section{Calibration map}
\label{sec:s-calib}

The conservative variant is called \emph{strict} in the experiment logs and configurations. \Cref{tab:s-calib} gives the actual vocabulary edits, extending the main paper's aggregate rule counts. Dropping a type removes every sighting of it. For Order questions, the number of undecidable cases attributable to dropped types rises from 7 to 19.

\begin{table}[h]
\centering
\caption{\textbf{Calibration rules} (fit on validation). $\star$: also in the conservative variant.}
\label{tab:s-calib}
\vspace{-0.6em}
\footnotesize
\begin{tabular}{@{}p{1.3cm}p{6.4cm}@{}}
\toprule
Rule & Types\\
\midrule
rename & Desk$^\star$, Table$^\star$, CoffeeTable$^\star$\,$\rightarrow$\,DiningTable\\
drop & Bathtub$^\star$, Cabinet$^\star$, Cup$^\star$, Lamp$^\star$, Mirror$^\star$, Ottoman$^\star$, SoapBar$^\star$, TableTopDecor$^\star$, TissueBox$^\star$, Apple, Bottle, DeskLamp, Fork, Knife, Laptop, Pen, Plate, RemoteControl, SideTable, SinkBasin, SoapBottle, Spoon, Stool, ToiletPaper, Vase\\
\bottomrule
\end{tabular}
\end{table}

\section{Room-classifier features and fitting}
\label{sec:s-rooms}

The simulator's leg index supplies the crossing events. For each object type, one feature is the fraction of a segment's frames containing that calibrated type. Additional features are the fractions receiving each of the four VLM room votes or no vote. Training labels are the majority true room within each segment. Multinomial logistic regression is fit on 17{,}664 training segments; inverse regularisation is selected from $\{0.1,0.3,1,3,10,30\}$ by validation segment accuracy. The naming ablation and fit statistics are in \cref{tab:s-roomfit}.

\section{Additional reader diagnostics}
\label{sec:s-reader}

\Cref{tab:s-undecided} separates further abstention rates from incorrect decisions. \Cref{tab:s-types} covers question types beyond the Order, Rooms-visited and Existence trade-offs quantified in the main paper. These are question-level breakdowns, rather than alternative definitions of answerability.

\para{Scope of the reader.}
The reader is hand-written for the benchmark's eight templated question types (main paper, Tab.~1). Answerability therefore measures retained evidence for these question forms only. Free-form questions such as those in OpenEQA~\cite{majumdar2024openeqa}, or questions needing multi-step reasoning, have no matching rule, and we have not tested how the idea transfers to them. A transfer would need either a rule set per question family or a reader whose own errors can be measured separately from missing evidence.

\begin{table}[h]
\centering
\caption{\textbf{Additional abstention rates.} The full repaired graph's rate is reported in the main paper.}
\label{tab:s-undecided}
\vspace{-0.6em}
\footnotesize
\begin{tabular}{@{}lc@{}}
\toprule
Graphs & No answer\\
\midrule
baseline & \cellcolor{hourPeach!27.28}49.6\%\\
\quad + vocabulary map & \cellcolor{hourPeach!25.85}47.0\%\\
\fixes{1+2} & \cellcolor{hourPeach!22.05}40.1\%\\
\quad + \fix{3} & \cellcolor{hourPeach!23.21}42.2\%\\
ours, conservative calibration & \cellcolor{hourPeach!22.99}41.8\%\\
perfect, agent's room & \cellcolor{hourPeach!13.86}25.2\%\\
perfect, physical room & \cellcolor{hourPeach!6.60}12.0\%\\
perfect, question strip & \cellcolor{hourPeach!0.61}1.1\%\\
\bottomrule
\end{tabular}
\end{table}

\begin{table*}[t]
\centering
\caption{\textbf{Additional question types} (test, 95\% CI). Ours, cons.: our graphs with conservative calibration. Perfect graphs file objects by the agent's room or by their physical room. The prior answers the most frequent correct value per type on training questions and uses no memory.}
\label{tab:s-types}
\vspace{-0.6em}
\footnotesize
\setlength{\tabcolsep}{4pt}
\begin{tabular}{@{}lrcccccc@{}}
\toprule
Type & $n$ & Baseline & Ours & Ours, cons. & Perfect, agent's & Perfect, physical & Prior\\
\midrule
Presence-change & 102 & \cellcolor{hourBlue!19.41}35.3\ci{26.0}{44.9} & \cellcolor{hourBlue!37.73}68.6\ci{59.5}{77.4} & \cellcolor{hourBlue!39.88}72.5\ci{63.7}{81.0} & \cellcolor{hourBlue!55.00}100\ci{100}{100} & \cellcolor{hourBlue!55.00}100\ci{100}{100} & \cellcolor{hourBlue!28.05}51.0\ci{41.2}{60.6}\\
Persistence & 131 & \cellcolor{hourBlue!14.30}26.0\ci{18.6}{33.6} & \cellcolor{hourBlue!24.36}44.3\ci{35.8}{52.7} & \cellcolor{hourBlue!26.89}48.9\ci{40.3}{57.4} & \cellcolor{hourBlue!28.55}51.9\ci{43.3}{60.7} & \cellcolor{hourBlue!49.12}89.3\ci{83.7}{94.4} & \cellcolor{hourBlue!40.31}73.3\ci{65.7}{80.6}\\
Room ID & 80 & \cellcolor{hourBlue!20.62}37.5\ci{27.0}{48.2} & \cellcolor{hourBlue!17.88}32.5\ci{22.5}{43.0} & \cellcolor{hourBlue!18.59}33.8\ci{23.7}{44.4} & \cellcolor{hourBlue!17.16}31.2\ci{21.2}{41.8} & \cellcolor{hourBlue!47.41}86.2\ci{78.2}{93.4} & \cellcolor{hourBlue!15.12}27.5\ci{17.9}{37.7}\\
Counting & 65 & \cellcolor{hourBlue!2.53}4.6\ci{0.0}{10.4} & \cellcolor{hourBlue!5.94}10.8\ci{3.8}{19.0} & \cellcolor{hourBlue!5.94}10.8\ci{3.8}{19.0} & \cellcolor{hourBlue!37.23}67.7\ci{55.9}{78.6} & \cellcolor{hourBlue!42.30}76.9\ci{66.2}{86.8} & \cellcolor{hourBlue!34.70}63.1\ci{50.8}{74.5}\\
Global count & 120 & \cellcolor{hourBlue!5.50}10.0\ci{5.0}{15.7} & \cellcolor{hourBlue!7.81}14.2\ci{8.2}{20.6} & \cellcolor{hourBlue!9.19}16.7\ci{10.3}{23.6} & \cellcolor{hourBlue!14.19}25.8\ci{18.2}{33.9} & \cellcolor{hourBlue!16.94}30.8\ci{22.7}{39.3} & \cellcolor{hourBlue!20.62}37.5\ci{28.7}{46.4}\\
\bottomrule
\end{tabular}
\vspace{-0.8em}
\end{table*}

\para{Exact question-prior values.}
The training modes are Order \emph{before}, Presence-change \emph{yes}, Rooms-visited \emph{4}, Persistence \emph{1}, Room ID \emph{bedroom}, Existence \emph{painting}, Counting \emph{2} and Global count \emph{4}. These are answer values rather than option positions. If the value is not among a question's options, the prior picks the most frequent correct letter for that type.

\section{Door-corruption implementation}
\label{sec:s-doors}

Corruptions are applied before segment formation. Timing perturbations shift each crossing uniformly in $[-k,k]$ frames; each new segment takes the original segment label covering most of it. Missed crossings merge segments and retain the longer part's label. Spurious crossings split a segment at a random interior frame, retain its label on both halves, and number the second as a new room. Crossing deletion uses probability $p$ per crossing; insertion uses probability $p$ per segment. Re-forming clean segments reproduces the uncorrupted construction.

\begin{table}[h]
\centering
\caption{\textbf{Additional crossing perturbations.} The main paper reports clean crossings, timing at eight frames, and the 10\% means. $\Delta$: paired difference to clean crossings, 95\% CI; n/a: no random corruption. RV: Rooms-visited answerability.}
\label{tab:s-doors}
\vspace{-0.6em}
\footnotesize
\setlength{\tabcolsep}{2.4pt}
\begin{tabular}{@{}lcccc@{}}
\toprule
Crossings & Answ. & Seed range & $\Delta$ & RV\\
\midrule
timing $k{=}1$ & \cellcolor{hourBlue!20.35}37.0 & 36.8--37.5 & \cellcolor{hourPeach!1.47}$-$0.40\ci{$-$0.81}{$-$0.01} & \cellcolor{hourBlue!32.73}59.5\\
timing $k{=}2$ & \cellcolor{hourBlue!20.35}37.0 & 36.6--37.6 & \cellcolor{hourPeach!1.50}$-$0.41\ci{$-$0.91}{+0.08} & \cellcolor{hourBlue!32.73}59.5\\
timing $k{=}4$ & \cellcolor{hourBlue!20.41}37.1 & 36.5--37.9 & \cellcolor{hourPeach!0.88}$-$0.24\ci{$-$0.91}{+0.43} & \cellcolor{hourBlue!32.78}59.6\\
missed $p{=}0.25$ & \cellcolor{hourBlue!18.81}34.2 & 32.6--35.1 & \cellcolor{hourPeach!11.77}$-$3.21\ci{$-$4.33}{$-$2.10} & \cellcolor{hourBlue!22.39}40.7\\
missed $p{=}0.50$ & \cellcolor{hourBlue!17.05}31.0 & 29.1--32.0 & \cellcolor{hourPeach!23.47}$-$6.40\ci{$-$8.21}{$-$4.62} & \cellcolor{hourBlue!14.52}26.4\\
spurious $p{=}0.25$ & \cellcolor{hourBlue!18.37}33.4 & 32.6--34.1 & \cellcolor{hourPeach!14.45}$-$3.94\ci{$-$5.15}{$-$2.75} & \cellcolor{hourBlue!17.76}32.3\\
spurious $p{=}0.50$ & \cellcolor{hourBlue!16.94}30.8 & 30.1--31.9 & \cellcolor{hourPeach!24.09}$-$6.57\ci{$-$8.21}{$-$5.01} & \cellcolor{hourBlue!7.75}14.1\\
\bottomrule
\end{tabular}
\vspace{-0.8em}
\end{table}

For the main paper's 10\% perturbations, paired changes are $-$1.29\ci{$-$1.86}{$-$0.72} for missed crossings and $-$1.82\ci{$-$2.50}{$-$1.19} for spurious crossings. Conservative calibration with per-frame rooms scores 37.8\ci{34.5}{41.0}.

\section{Detector threshold behaviour}
\label{sec:s-gdino}

Prompting eight classes together reduced Grounding DINO's maximum score to 0.12, versus 0.30 for one class. Per-class thresholds were fit on 300 validation frames from 30 episodes. Validation F1 was 55.7, and 18 classes were switched off entirely, indicating overfit. A global threshold of 0.5 instead yielded F1 28.9. On the 300-frame pilot, Grounding DINO~\cite{liu2024groundingdino} with the validation-fit thresholds reached F1 45.9 versus 52.6 for MomaGraph-R1 with the closed prompt, at 10.7 versus about 0.65\,s/frame. This comparison was run after the repairs were chosen. OWLv2~\cite{minderer2023owlv2} was not tested. \Cref{tab:s-gdino} adds precision and recall.

\begin{table}[h]
\centering
\caption{\textbf{Detector precision and recall} (300 test frames, type level).}
\label{tab:s-gdino}
\vspace{-0.6em}
\footnotesize
\setlength{\tabcolsep}{3.5pt}
\begin{tabular}{@{}lcc@{}}
\toprule
Model & P & R\\
\midrule
MomaGraph-R1, closed prompt & \cellcolor{hourBlue!27.89}50.7 & \cellcolor{hourBlue!29.98}54.5\\
Grounding DINO, per-class thr. & \cellcolor{hourBlue!24.20}44.0 & \cellcolor{hourBlue!26.45}48.1\\
Grounding DINO, global thr.\ 0.5 & \cellcolor{hourBlue!11.05}20.1 & \cellcolor{hourBlue!28.38}51.6\\
\bottomrule
\end{tabular}
\vspace{-0.8em}
\end{table}

\section{Optimisation details}
\label{sec:s-train}

LoRA updates 88.8M of 8.38B parameters, including the vision encoder. Optimisation uses AdamW with learning rate $10^{-4}$, cosine decay to 5\% of peak, 5\% warm-up, weight decay 0.01 and gradient clipping at 1.0. The micro-batch is one; arithmetic is bfloat16 and each run uses one H200 GPU. The navigation prompt in \cref{sec:s-prompts} supplies the auxiliary next-action classifier. Navigation performance is not evaluated.

\section{Real-video memory construction}
\label{sec:s-vsi}

Frames are spaced evenly through each scan and resized on the longest side. The text is inserted before the question as \emph{Memory of what you saw}. Since scans lack room labels, the renderer omits rooms and lists per-frame objects together with first/last frame indices and maximum single-frame multiplicity. Each scene's other-scene control is paired with a different scan. Despite the closed prompt, 37\% of emitted names are outside the question vocabulary. The per-task results beyond the main paper's counting and overall rows are in \cref{tab:s-vsi}.

\para{Why correct and unrelated memory score alike.}
The answerer sees the same 16 frames in every frame-based condition, so memory text is the only difference. Correct memory scores 44.0 and another scene's memory 43.8, both below frames alone (47.3). If the answerer drew on scene-specific graph content, correct memory should beat the unrelated control. That it does not suggests the loss comes from the presence of the memory text. Three causes fit this pattern and our runs cannot separate them. (i) The added text lengthens the input and may dilute attention to the frames. (ii) Memory entries can conflict with the frames: 37\% of emitted names fall outside the question vocabulary, and the per-frame maximum count can undercount objects seen across frames, which is consistent with Counting falling from 46.5 to 35.3. (iii) The answerer may not use graph content at all. Memory alone scores 28.6, so the graphs carry some question-relevant content, but this run does not show whether the answerer reads it.

\para{Scope.}
This check uses one graph extractor and one answerer, and it omits calibration and door segmentation because scans have no room labels or door events. It tests the graph-to-answerer interface under these settings and does not evaluate the full repaired pipeline. Separating construction from use would need input-matched controls: filler text of the same length, the same graph with entries shuffled across objects, and a per-question check of whether the graph contains the answering evidence.

\section{Additional experiment records}
\label{sec:s-logs}

\Cref{tab:s-files} specifies the experiment bundle intended for anonymised review. It identifies the configurations, predictions and scripts needed to reproduce the results. The tables and plots below reproduce selected records; raw experiment assets are separate from the manuscript sources.

\begin{table*}[t]
\centering
\caption{\textbf{Experiment-bundle specification.} Named main-paper results refer to their tables there.}
\label{tab:s-files}
\vspace{-0.6em}
\footnotesize
\setlength{\tabcolsep}{4pt}
\begin{tabular}{@{}>{\raggedright\arraybackslash}p{2.9cm}>{\raggedright\arraybackslash}p{9.0cm}>{\raggedright\arraybackslash}p{4.8cm}@{}}
\toprule
Folder & Contents & Supports\\
\midrule
\texttt{test\_predictions/} & correctness and question type of each of the 800 test questions, for every QA run, control and test-time swap & Main QA results; \cref{tab:s-testtime}\\
\texttt{training\_logs/} & arguments, losses, gradient norms and validation accuracy every 500 steps of the eight QA runs & \cref{tab:s-runs}, \cref{fig:s-val}\\
\texttt{kg\_evaluation/} & answerability per question for every graph set and the prior; frame and episode scores; room confusion; detector pilot & Main memory results; \cref{tab:s-undecided,tab:s-types,tab:s-gdino}\\
\texttt{scene\_graphs\_test/} & test-split graphs: baseline, \fixes{1+2}, +\fix{3}, ours, conservative & Main repair ladder; \cref{fig:qual,fig:failures}\\
\texttt{configs/} & calibration maps, vocabulary map, the prompt of each graph set & \cref{sec:s-prompts,sec:s-calib}\\
\texttt{pipeline\_logs/} & prompt and resolution pilot, calibration fit, room-classifier fit, false positives by type & \cref{tab:s-pilot,tab:s-fp,tab:s-roomfit}, \cref{fig:s-calib}\\
\texttt{scripts/} & reader, calibration and room fitting, re-rendering, graph extraction, bootstrap comparisons, launch scripts & all\\
\texttt{vsibench/} & per-question outputs of every condition and backbone, scripts, logs & Main real-video results; \cref{tab:s-vsi}\\
\bottomrule
\end{tabular}
\vspace{-0.8em}
\end{table*}

\subsection{Training trajectories}
\begin{table}[h]
\centering
\caption{\textbf{Validation trajectories}, from the training logs. Validation: answer accuracy (\%) on 400 validation questions at steps 1k, 3k and 6k; best: highest validation accuracy and its step. Every run completed 6{,}000 optimiser steps with no skipped step; the test score always uses step 6k.}
\label{tab:s-runs}
\vspace{-0.6em}
\footnotesize
\setlength{\tabcolsep}{2.6pt}
\begin{tabular}{@{}lcccccc@{}}
\toprule
Memory & Seed & Started & Val 1k & Val 3k & Val 6k & Best (step)\\
\midrule
none & 0 & 15 Sep & \cellcolor{hourBlue!34.65}63.0 & \cellcolor{hourBlue!37.95}69.0 & \cellcolor{hourBlue!37.02}67.3 & \cellcolor{hourBlue!38.39}69.8 (5k)\\
none & 1 & 17 Sep & \cellcolor{hourBlue!36.85}67.0 & \cellcolor{hourBlue!36.85}67.0 & \cellcolor{hourBlue!37.29}67.8 & \cellcolor{hourBlue!37.56}68.3 (2.5k)\\
baseline graphs & 0 & 16 Sep & \cellcolor{hourBlue!36.58}66.5 & \cellcolor{hourBlue!39.33}71.5 & \cellcolor{hourBlue!41.25}75.0 & \cellcolor{hourBlue!41.25}75.0 (6k)\\
baseline graphs & 1 & 17 Sep & \cellcolor{hourBlue!36.85}67.0 & \cellcolor{hourBlue!39.05}71.0 & \cellcolor{hourBlue!39.77}72.3 & \cellcolor{hourBlue!40.15}73.0 (4k)\\
ours & 0 & 1 Oct & \cellcolor{hourBlue!41.25}75.0 & \cellcolor{hourBlue!42.62}77.5 & \cellcolor{hourBlue!43.45}79.0 & \cellcolor{hourBlue!44.16}80.3 (5.5k)\\
ours & 1 & 1 Oct & \cellcolor{hourBlue!40.42}73.5 & \cellcolor{hourBlue!42.24}76.8 & \cellcolor{hourBlue!44.00}80.0 & \cellcolor{hourBlue!44.00}80.0 (6k)\\
perfect graph & 0 & 28 Sep & \cellcolor{hourBlue!44.33}80.6 & \cellcolor{hourBlue!45.38}82.5 & \cellcolor{hourBlue!47.47}86.3 & \cellcolor{hourBlue!48.12}87.5 (5k)\\
perfect graph & 1 & 14 Oct & \cellcolor{hourBlue!47.08}85.6 & \cellcolor{hourBlue!48.45}88.1 & \cellcolor{hourBlue!47.47}86.3 & \cellcolor{hourBlue!48.45}88.1 (2k)\\
\bottomrule
\end{tabular}
\vspace{-0.8em}
\end{table}

\begin{figure}[h]
\centering
\includegraphics[width=\linewidth]{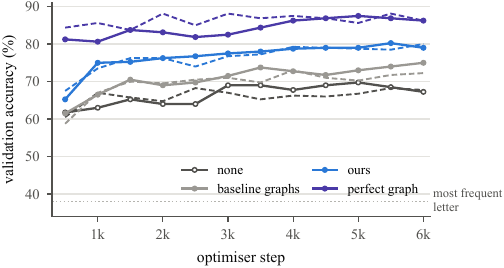}
\vspace{-1.6em}
\caption{\textbf{Validation accuracy during QA training} (400 validation questions, every 500 steps). Solid: seed 0; dashed: seed 1. Dotted: the share of the most frequent correct letter.}
\label{fig:s-val}
\vspace{-0.6em}
\end{figure}

The repaired-memory models lead their baseline counterparts at all validation passes. No run's most frequent predicted letter exceeds 49\% of validation answers. Initial starts of the no-memory and baseline seed-0 runs were stopped before their first validation pass and restarted with identical arguments. The no-memory seeds answer 514 and 523 test questions correctly; the perfect-memory seeds answer 663 and 672. Both repaired-memory seeds improve on baseline memory: 608 and 606 correct, versus 572 and 571.

\subsection{Graph-generation and fitting records}
\begin{table}[h]
\centering
\caption{\textbf{Prompt and resolution pilot} (MomaGraph-R1, the same 300 test frames from 30 episodes, type level). Only the prompt and the input image change. P/R: precision/recall (\%); objects: types listed per frame; s: seconds per frame on one GPU. Rows: prompt (open = the baseline's, closed = \fix{2}), input size in px. n/a: stored graphs are not timed. Shading throughout: blue for scores, peach for unanswered/absent rates (0--100\%); signed changes use blue for gains and peach for losses ($\pm15$\,pp).}
\label{tab:s-pilot}
\vspace{-0.6em}
\footnotesize
\setlength{\tabcolsep}{2.6pt}
\begin{tabular}{@{}lcccccc@{}}
\toprule
Prompt, input & P & R & F1 & Objects & Room & s\\
\midrule
open, 256, stored graphs & \cellcolor{hourBlue!25.68}46.7 & \cellcolor{hourBlue!13.80}25.1 & \cellcolor{hourBlue!17.93}32.6 & 2.48 & \cellcolor{hourBlue!26.79}48.7 & n/a\\
open, 256, re-run & \cellcolor{hourBlue!25.85}47.0 & \cellcolor{hourBlue!14.41}26.2 & \cellcolor{hourBlue!18.48}33.6 & 2.57 & \cellcolor{hourBlue!25.68}46.7 & 0.57\\
closed, 256 & \cellcolor{hourBlue!21.12}38.4 & \cellcolor{hourBlue!23.21}42.2 & \cellcolor{hourBlue!22.11}40.2 & 5.07 & \cellcolor{hourBlue!25.46}46.3 & 0.69\\
closed, 256 upscaled to 512 & \cellcolor{hourBlue!24.53}44.6 & \cellcolor{hourBlue!26.84}48.8 & \cellcolor{hourBlue!25.63}46.6 & 5.05 & \cellcolor{hourBlue!26.23}47.7 & 0.66\\
closed, 512 (\fixes{1+2}) & \cellcolor{hourBlue!27.89}50.7 & \cellcolor{hourBlue!29.98}54.5 & \cellcolor{hourBlue!28.93}52.6 & \multicolumn{3}{c}{not logged}\\
\bottomrule
\end{tabular}
\vspace{-0.8em}
\end{table}

A rerun of the original prompt reproduces stored-graph F1 closely (33.6 versus 32.6). Upscaling the same image adds 6.6 recall points without new detail; re-rendering adds another 5.7. This distinguishes model input size from newly resolved visual detail.

\begin{figure*}[t]
\centering
\includegraphics[width=\textwidth]{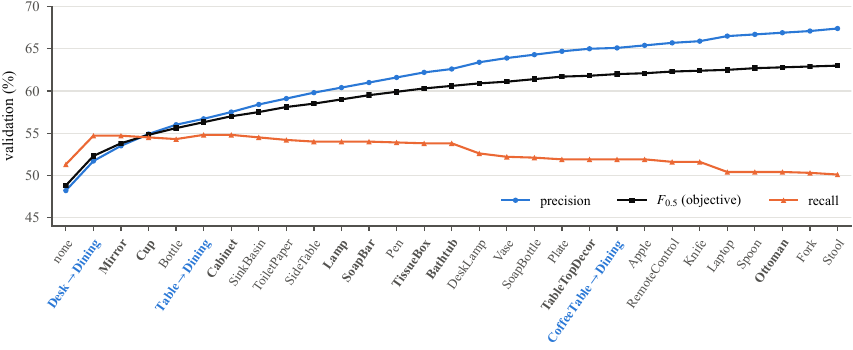}
\vspace{-1.4em}
\caption{\textbf{Greedy calibration fit} (\fix{3}, validation, type level). Each step adds the rule that most raises $F_{0.5}$; rules are listed in the order they were added. Blue: renames to DiningTable; bold: rules kept in the conservative variant.}
\label{fig:s-calib}
\vspace{-0.6em}
\end{figure*}

Validation $F_{0.5}$ rises from 48.8 to 63.0 during greedy fitting, with precision moving from 48.2 to 67.4 and recall from 51.3 to 50.1. The first five rules provide half the gain. \Cref{tab:s-fp} resolves which categories drive false positives; \cref{tab:s-roomfit} records the room-classifier fit.

\begin{table*}[t]
\centering
\caption{\textbf{False positives of the \fixes{1+2} graphs} (test, all frames, type level): the 22 types with the most false positives, which together make 75\% of all 21{,}117 false positives. Absent: share of the type's false positives in episodes where the type never appears. Rule: what \fix{3} learned on validation ($\star$: kept in the conservative variant).}
\label{tab:s-fp}
\vspace{-0.6em}
\footnotesize
\setlength{\tabcolsep}{3pt}
\begin{tabular}{@{}lrccl@{\hspace{28pt}}lrccl@{}}
\toprule
Type & FP & Prec. & Absent & Rule & Type & FP & Prec. & Absent & Rule\\
\midrule
Chair & 2{,}534 & \cellcolor{hourBlue!23.87}43.4 & \cellcolor{hourPeach!13.20}24\% & keep & Painting & 521 & \cellcolor{hourBlue!41.97}76.3 & \cellcolor{hourPeach!6.05}11\% & keep\\
Desk & 2{,}339 & \cellcolor{hourBlue!4.07}7.4 & \cellcolor{hourPeach!50.60}92\% & rename$^\star$ & Bed & 497 & \cellcolor{hourBlue!42.41}77.1 & \cellcolor{hourPeach!4.95}9\% & keep\\
Cup & 906 & \cellcolor{hourBlue!1.98}3.6 & \cellcolor{hourPeach!51.70}94\% & drop$^\star$ & SideTable & 495 & \cellcolor{hourBlue!8.80}16.0 & \cellcolor{hourPeach!41.80}76\% & drop\\
DeskLamp & 861 & \cellcolor{hourBlue!14.30}26.0 & \cellcolor{hourPeach!35.75}65\% & drop & Laptop & 471 & \cellcolor{hourBlue!24.97}45.4 & \cellcolor{hourPeach!23.10}42\% & drop\\
Bottle & 835 & \cellcolor{hourBlue!4.34}7.9 & \cellcolor{hourPeach!45.65}83\% & drop & Sink & 456 & \cellcolor{hourBlue!32.84}59.7 & \cellcolor{hourPeach!20.35}37\% & keep\\
Mirror & 779 & \cellcolor{hourBlue!0.00}0.0 & \cellcolor{hourPeach!55.00}100\% & drop$^\star$ & Lamp & 390 & \cellcolor{hourBlue!0.00}0.0 & \cellcolor{hourPeach!55.00}100\% & drop$^\star$\\
Television & 663 & \cellcolor{hourBlue!33.55}61.0 & \cellcolor{hourPeach!8.25}15\% & keep & Pillow & 378 & \cellcolor{hourBlue!35.42}64.4 & \cellcolor{hourPeach!9.35}17\% & keep\\
Cabinet & 656 & \cellcolor{hourBlue!11.66}21.2 & \cellcolor{hourPeach!37.95}69\% & drop$^\star$ & Faucet & 365 & \cellcolor{hourBlue!33.44}60.8 & \cellcolor{hourPeach!18.70}34\% & keep\\
ToiletPaper & 559 & \cellcolor{hourBlue!6.71}12.2 & \cellcolor{hourPeach!32.45}59\% & drop & Table & 363 & \cellcolor{hourBlue!0.00}0.0 & \cellcolor{hourPeach!55.00}100\% & rename$^\star$\\
Vase & 557 & \cellcolor{hourBlue!21.34}38.8 & \cellcolor{hourPeach!18.15}33\% & drop & TVStand & 363 & \cellcolor{hourBlue!33.22}60.4 & \cellcolor{hourPeach!23.10}42\% & keep\\
SinkBasin & 546 & \cellcolor{hourBlue!8.09}14.7 & \cellcolor{hourPeach!35.20}64\% & drop & SoapBar & 351 & \cellcolor{hourBlue!2.53}4.6 & \cellcolor{hourPeach!51.70}94\% & drop$^\star$\\
\bottomrule
\end{tabular}
\vspace{-0.8em}
\end{table*}

\begin{table}[h]
\centering
\caption{\textbf{Room inference} on all frames of the validation (500 episodes) and test (200) splits. Segments are the frames between two door crossings. The classifier is trained on 17{,}664 training segments; validation segment accuracy is 74.9--76.2\% over the regularisation grid, and $C{=}10$ is chosen. The true route switches rooms 2.1 times per episode in both splits.}
\label{tab:s-roomfit}
\vspace{-0.6em}
\footnotesize
\setlength{\tabcolsep}{2.4pt}
\begin{tabular}{@{}lcccccc@{}}
\toprule
& \multicolumn{3}{c}{Validation} & \multicolumn{3}{c}{Test}\\
\cmidrule(lr){2-4}\cmidrule(l){5-7}
Rooms from & Acc. & Switch. & Exact & Acc. & Switch. & Exact\\
\midrule
per-frame VLM & \cellcolor{hourBlue!28.11}51.1 & 13.7 & \cellcolor{hourBlue!19.91}181/500 & \cellcolor{hourBlue!27.28}49.6 & 12.8 & \cellcolor{hourBlue!19.80}72/200\\
segment, VLM vote & \cellcolor{hourBlue!35.31}64.2 & 1.7 & \cellcolor{hourBlue!54.56}496/500 & \cellcolor{hourBlue!33.83}61.5 & 1.5 & \cellcolor{hourBlue!54.73}199/200\\
segment, classifier & \cellcolor{hourBlue!41.52}75.5 & 1.6 & \cellcolor{hourBlue!54.56}496/500 & \cellcolor{hourBlue!39.71}72.2 & 1.5 & \cellcolor{hourBlue!54.73}199/200\\
\bottomrule
\end{tabular}
\vspace{-0.8em}
\end{table}

\subsection{Memory substitutions without retraining}
\begin{table}[h]
\centering
\caption{\textbf{Changing the memory at test time only.} The model trained on baseline graphs (two seeds) is evaluated with other memory; no retraining. $\Delta$: seed-averaged paired difference to its own memory, 95\% CI (bold: excludes zero). $^\dagger$Diagnostic: false positives (FP) removed using simulator truth. $^\ddagger$Only that part of the predicted graph is replaced by the simulator's.}
\label{tab:s-testtime}
\vspace{-0.6em}
\footnotesize
\setlength{\tabcolsep}{2.2pt}
\begin{tabular}{@{}lcccc@{}}
\toprule
Memory at test time & Seed 0 & Seed 1 & Mean & $\Delta$\\
\midrule
\midrule
\quad + vocabulary map & 586 & 566 & 576.0 & \cellcolor{hourBlue!2.05}+0.56\ci{$-$0.75}{+1.88}\\
\fixes{1+2} & 576 & 573 & 574.5 & \cellcolor{hourBlue!1.39}+0.38\ci{$-$1.62}{+2.38}\\
\quad FP removed$^\dagger$ & 590 & 585 & 587.5 & \cellcolor{hourBlue!7.33}+2.00\ci{+0.00}{+4.00}\\
baseline, FP removed$^\dagger$ & 564 & 560 & 562.0 & \cellcolor{hourPeach!4.36}$-$1.19\ci{$-$3.44}{+1.06}\\
\midrule
true relations$^\ddagger$ & 573 & 567 & 570.0 & \cellcolor{hourPeach!0.70}$-$0.19\ci{$-$1.44}{+1.06}\\
true rooms$^\ddagger$ & 572 & 581 & 576.5 & \cellcolor{hourBlue!2.27}+0.62\ci{$-$1.25}{+2.50}\\
true objects$^\ddagger$ & 586 & 594 & 590.0 & \cellcolor{hourBlue!8.47}\textbf{+2.31}\ci{+0.31}{+4.31}\\
true objects + rooms$^\ddagger$ & 591 & 600 & 595.5 & \cellcolor{hourBlue!11.00}\textbf{+3.00}\ci{+0.62}{+5.44}\\
perfect graph & 590 & 597 & 593.5 & \cellcolor{hourBlue!10.08}\textbf{+2.75}\ci{+0.50}{+5.00}\\
\bottomrule
\end{tabular}
\vspace{-0.8em}
\end{table}

The substitutions preceded calibration. Removing false positives from the higher-resolution closed-prompt graphs improves QA by 2.00 points\ci{+0.00}{+4.00}; this diagnostic motivated calibration. With seed 0 alone, the perfect-graph swap gives +2.25\ci{$-$0.25}{+4.75}. Additional content/structure controls are in \cref{tab:s-perturb}.

\subsection{Other real-video tasks and backbones}
\begin{table}[h]
\centering
\caption{\textbf{Additional VSI-Bench tasks}, Qwen3-VL-8B, official scores. Overall averages eight task scores, with the three relative-direction tasks merged into one. Frames: frames only; + mem.: frames and our scene-graph memory; + other: frames and another scene's memory. Bold: best per row.}
\label{tab:s-vsi}
\vspace{-0.6em}
\footnotesize
\setlength{\tabcolsep}{2.6pt}
\begin{tabular}{@{}lrcccc@{}}
\toprule
Task & $n$ & Frames & + mem. & + other & Mem. only\\
\midrule
Absolute distance & 434 & \cellcolor{hourBlue!20.41}\textbf{37.1} & \cellcolor{hourBlue!20.02}36.4 & \cellcolor{hourBlue!19.96}36.3 & \cellcolor{hourBlue!9.68}17.6\\
Appearance order & 318 & \cellcolor{hourBlue!32.51}\textbf{59.1} & \cellcolor{hourBlue!29.59}53.8 & \cellcolor{hourBlue!25.41}46.2 & \cellcolor{hourBlue!23.38}42.5\\
Object size & 353 & \cellcolor{hourBlue!33.77}\textbf{61.4} & \cellcolor{hourBlue!32.06}58.3 & \cellcolor{hourBlue!31.84}57.9 & \cellcolor{hourBlue!17.71}32.2\\
Rel.\ dir., easy & 212 & \cellcolor{hourBlue!31.13}56.6 & \cellcolor{hourBlue!31.41}\textbf{57.1} & \cellcolor{hourBlue!30.09}54.7 & \cellcolor{hourBlue!22.05}40.1\\
Rel.\ dir., hard & 116 & \cellcolor{hourBlue!15.62}28.4 & \cellcolor{hourBlue!15.62}28.4 & \cellcolor{hourBlue!14.69}26.7 & \cellcolor{hourBlue!16.61}\textbf{30.2}\\
Rel.\ dir., medium & 54 & \cellcolor{hourBlue!26.45}\textbf{48.1} & \cellcolor{hourBlue!23.43}42.6 & \cellcolor{hourBlue!25.46}46.3 & \cellcolor{hourBlue!17.32}31.5\\
Relative distance & 310 & \cellcolor{hourBlue!27.66}50.3 & \cellcolor{hourBlue!26.95}49.0 & \cellcolor{hourBlue!27.83}\textbf{50.6} & \cellcolor{hourBlue!19.36}35.2\\
Room size & 200 & \cellcolor{hourBlue!30.69}\textbf{55.8} & \cellcolor{hourBlue!29.37}53.4 & \cellcolor{hourBlue!28.71}52.2 & \cellcolor{hourBlue!18.37}33.4\\
Route plan & 114 & \cellcolor{hourBlue!13.04}23.7 & \cellcolor{hourBlue!12.54}22.8 & \cellcolor{hourBlue!14.96}\textbf{27.2} & \cellcolor{hourBlue!12.04}21.9\\
\midrule
\bottomrule
\end{tabular}
\vspace{-0.8em}
\end{table}

With identical frame inputs and decoding, MomaGraph-R1 scores 26.3 overall and its base Qwen2.5-VL-7B scores 26.2; InternVL3.5-8B scores 47.5. These backbone checks motivated the choice of answering model. Task-specific changes include $-$5.6 for medium relative direction and $-$5.3 for appearance order.



\section{Development details and visibility checks}
\label{sec:s-faq}

\para{Which choices saw test data.}
Two development steps used test data. First, the 300-frame prompt and resolution pilot (\supp{Tab.~S8}) sampled test frames and ran while \fixes{1+2} were being chosen. Second, the conservative-calibration tie-break below was made with test numbers visible. All other fitting used training or validation data only: the full calibration map (validation $F_{0.5}$), the room classifier (training fit, validation-selected regularisation) and the reader rules, which have no fitted parameters. The paired bootstrap intervals describe variation over test episodes; they do not account for the choice of configurations, so the main-paper comparisons are exploratory. A confirmatory test would repeat the pilot on validation frames, freeze every choice, and score the frozen pipeline once on houses unused so far.

\para{Conservative-selection tie.}
Conservative calibration and renames-only tied at 36.5 validation answerability. The comparison printed validation and test numbers together; the tie was broken by validation precision while test results were visible. Renames-only scores 37.2 on test and conservative calibration 37.8. Conservative selection followed the starts of both full-calibration retraining runs, each taking about a day, so no conservative QA run was added.

\para{Small objects needed by questions.}
Visibility begins at about $6{\times}6$ pixels in a stored frame. Only 31 test questions hinge on a small object never seen from closer than 3\,m, distinguishing missed near-field evidence from questions requiring distant detail.

\para{Provenance.}
The experiment-order table distinguishes the inherited runs identified in the main paper from the subsequent graph diagnostics, simulator interventions, repairs and real-video tests.

\section{Benchmark and dataset details}
\label{sec:s-examples}

This section adds statistics and examples to the main paper's benchmark and dataset description. Test statistics are computed from the test episode records.

\begin{figure}[t]
\centering
\includegraphics[width=0.32\linewidth]{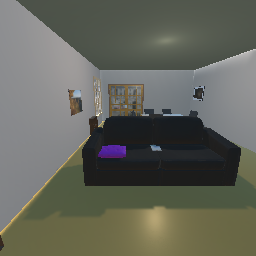}\hfill
\includegraphics[width=0.32\linewidth]{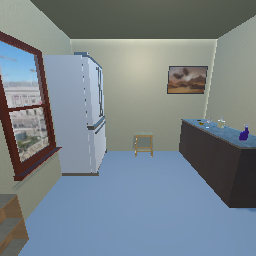}\hfill
\includegraphics[width=0.32\linewidth]{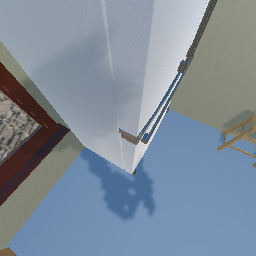}
\vspace{-0.4em}
\caption{\textbf{Stored $256{\times}256$ frames from one episode} outside the test split (31 frames, living room\,$\rightarrow$\,kitchen). Left: an early view in the living room. Middle: the view at the room transition, with the kitchen ahead. Right: a late view near the end of the route.}
\label{fig:s-example}
\vspace{-0.8em}
\end{figure}

\para{Episodes.}
The reference route starts out of view of the target object instance and leads to it. The actions are MoveAhead by 0.25, 0.5 or 1.0\,m, RotateLeft/RotateRight by 15, 30, 45 or 90$^\circ$, LookUp/LookDown by 15 or 30$^\circ$ within a pitch range of $-30$ to 60$^\circ$, and Done. Each step also stores the agent pose. All 200 test routes reach their target with these discrete actions alone. \Cref{fig:s-example} shows views from one episode outside the test split.

\para{Visible-object record.}
An object enters a frame's record when AI2-THOR flags it visible within 6\,m and its visible region covers at least 0.05\% of the image (about 33 pixels), so partly occluded and cut-off objects count. The room in each entry is the room the object stands in, which can differ from the agent's room when the object is seen through a doorway. Test frames contain 6.1 visible objects on average (median 5). A test episode shows 46.5 distinct object instances of 26.5 types on average, and 94 of the 106 types in the closed prompt (\fix{2}) appear in test frames.

\begin{figure}[t]
\centering
\includegraphics[width=\linewidth]{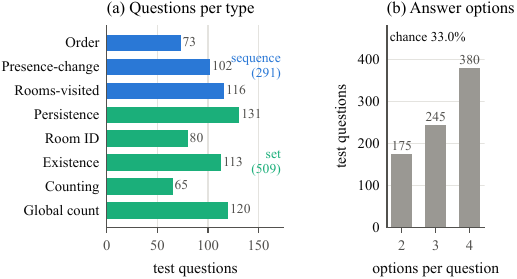}
\vspace{-1.5em}
\caption{\textbf{Test-split composition.} \textbf{(a)} Questions per type. \textbf{(b)} Options per question; the mix of two, three and four options puts uniform-guess accuracy at 33.0\%, rather than the 25\% of four options.}
\label{fig:s-dataset}
\vspace{-0.6em}
\end{figure}

\para{Questions and answers.}
No language model writes questions or answers. Every gold answer is computed from the visible-object records and room labels of the question strip (\cref{tab:qtypes}). Answers therefore follow the strip even when a question's wording refers to the whole run or house. Each question cites one to four strip frames, and its options are shuffled. The test split has 175 two-option, 245 three-option and 380 four-option questions (\cref{fig:s-dataset}b).

\para{Splits and route lengths.}
The houses come from ProcTHOR-10k's own training, validation and test partitions. A test house contributes one to five episodes (33 of the 53 contribute five). Each route stratum holds 50 test episodes (\cref{tab:s-strata}). Room labels name room types, so the 62 test routes that pass two rooms of the same type, including all of S4, contain two rooms with one label. Across test frames, 44\% are in bedrooms and 17--21\% in each of the other three room types; the target room is a bathroom in 89 of the 200 test episodes.

\begin{table}[h]
\centering
\caption{\textbf{Test split by route stratum.} Doors: door crossings on the route. Rooms: distinct rooms on the route. Frames: mean (range). Path: walked length (m). Objects: distinct object instances seen per episode. Repeat: episodes passing two rooms of the same type.}
\label{tab:s-strata}
\vspace{-0.6em}
\footnotesize
\setlength{\tabcolsep}{3.2pt}
\begin{tabular}{@{}lrccrrrr@{}}
\toprule
Stratum & Eps. & Doors & Rooms & Frames & Path & Objects & Repeat\\
\midrule
S1 & 50 & 1 & 2 & 20.4 (10--40) & 6.6 & 28.0 & 0\\
S2 & 50 & 2 & 2--3 & 33.6 (18--58) & 11.2 & 43.5 & 2\\
S3 & 50 & 3 & 4 & 46.4 (26--87) & 16.1 & 46.2 & 10\\
S4 & 50 & 4--6 & 5--7 & 66.2 (34--116) & 22.9 & 68.2 & 50\\
\bottomrule
\end{tabular}
\vspace{-0.8em}
\end{table}

\begin{table*}[t]
\centering
\caption{\textbf{Question types, gold answers and examples.} Options are shown in brackets. Every gold answer is computed from the visible-object records and room labels of the question strip. The main paper gives the reader's rules; this table adds the gold-answer definitions, the question wording and the test distribution.}
\label{tab:qtypes}
\vspace{-0.6em}
\footnotesize
\setlength{\tabcolsep}{4pt}
\renewcommand{\arraystretch}{1.08}
\begin{tabular}{@{}>{\raggedright\arraybackslash}p{1.95cm}r>{\raggedright\arraybackslash}p{5.3cm}>{\raggedright\arraybackslash}p{8.55cm}@{}}
\toprule
Type & $n$ & Gold answer & Example test question\\
\midrule
\multicolumn{4}{@{}l}{\emph{Sequence questions: when things were seen} (291)}\\
Order & 73 & whether type $X$ is first seen before or after type $Y$ & Did the robot first see a sink before or after it first saw a houseplant? [before, after]\\
Presence-change & 102 & whether a type seen in an earlier strip frame is absent from the strip's last frame & Was a statue visible at some point but then gone from view later? [yes, no]\\
Rooms-visited & 116 & number of distinct room labels among the strip frames & How many different rooms did the robot enter during its run? [2, 3, 4, 5]\\
\midrule
\multicolumn{4}{@{}l}{\emph{Set questions: what was where} (509)}\\
Persistence & 131 & number of distinct room labels whose frames show the type & In how many different rooms did the robot see a painting? [2, 3, 4]\\
Room ID & 80 & room label of the frames that show the type & In which room did the robot see a bed? [LivingRoom, Bedroom, Bathroom]\\
Existence & 113 & the option type seen in frames of the named room & Which of these did the robot see in the Bathroom? [desklamp, plunger, statue, pencil]\\
Counting & 65 & distinct instance IDs of the type in frames of the named room & How many distinct chair objects did the robot see in the Kitchen? [1, 2, 3]\\
Global count & 120 & distinct instance IDs of a category (\eg storage units) over the strip & In total, how many distinct electronic objects did the robot see anywhere in the house? [3, 4, 5]\\
\bottomrule
\end{tabular}
\vspace{-0.8em}
\end{table*}

\section{Visual examples and additional error measurements}
\label{sec:s-visual}

The frame-level illustrations expand the main paper's teaser episode by showing raw predicted names before normalization. The aggregate panels resolve object type, distance and room confusion.

\begin{figure}[t]
\centering
\includegraphics[width=\linewidth]{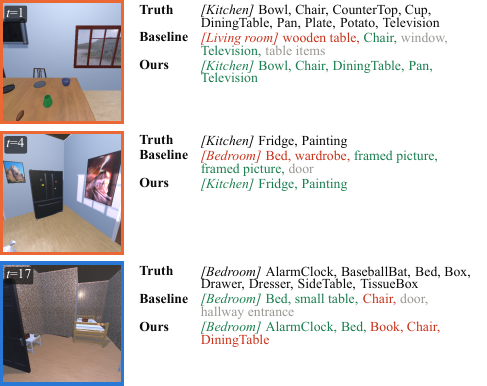}
\vspace{-1.5em}
\caption{\textbf{Three frames of the teaser episode, as each memory records them.} \emph{Truth}: visible object types and room. \emph{Baseline}: the VLM's raw names, coloured by what the renderer makes of them (green: a visible type; red: a type not in view; grey: dropped as architecture or outside the vocabulary). Room tags are green when correct. At $t{=}1$ the baseline calls the dining table a ``wooden table'' (mapped to Table, the wrong level of detail) and the kitchen a living room; at $t{=}4$ it places a bedroom inside the kitchen. Ours fixes both of these errors, but at $t{=}17$ it still adds furniture that is not there (Book, Chair, DiningTable).}
\label{fig:qual}
\vspace{-0.8em}
\end{figure}

\begin{figure*}[t]
\centering
\includegraphics[width=\textwidth]{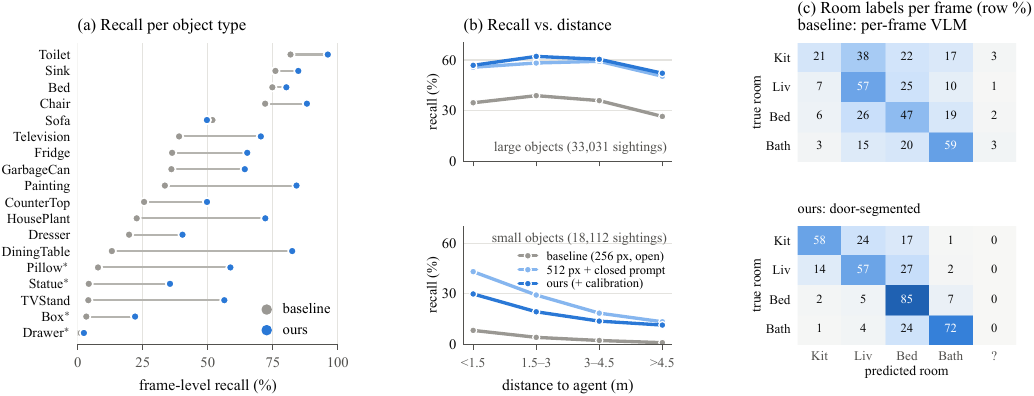}
\vspace{-1.6em}
\caption{\textbf{Where the baseline graphs fail, and what the fixes recover} (test, all 8{,}335 frames). \textbf{(a)} Recall of the 18 most frequent object types, baseline (grey) \vs ours (blue); $^\ast$\,marks small objects. Recall rises for 17 of the 18 types, most for small objects and for furniture the baseline names at the wrong level of detail (DiningTable 13\%\,$\rightarrow$\,83\%). \textbf{(b)} Recall by distance from the agent, for large and small objects. Higher resolution and the closed prompt (\fixes{1+2}, light blue) lift every distance bin; calibration (\fix{3}) gives up some small-object recall for precision. \textbf{(c)} Room labels per frame as row-normalised confusion matrices (``?'' = no room). The per-frame VLM confuses kitchens with living rooms, while door-segmented rooms (\fix{4}) roughly halve the errors and switch rooms 1.5 times per episode (true: 2.1).}
\label{fig:failures}
\vspace{-0.6em}
\end{figure*}

\begin{table}[t]
\centering
\caption{\textbf{Additional baseline diagnostics}, beyond the recall, hallucination and room-switch summaries in the main paper.}
\label{tab:s-extra-failures}
\footnotesize
\begin{tabular}{@{}>{\raggedright\arraybackslash}p{3.6cm}>{\raggedright\arraybackslash}p{4.2cm}@{}}
\toprule
Diagnostic & Measurement\\
\midrule
False positives by name & Table 1{,}182; SideTable 1{,}174\\
Architecture mentions & 35\% of mentions\\
Names outside vocabulary & 44\% of object names\\
Distinct-object count error & 1.30; perfect type-only graph 0.35\\
\bottomrule
\end{tabular}
\end{table}

Small objects account for 35\% of sightings. The benchmark vocabulary covers 94.5\% of ground-truth mentions. Restricting correctness to objects physically within the agent's room lowers baseline precision from 46.0\% to 30.4\%, quantifying how often visible objects belong to neighbouring rooms. Among the higher-resolution closed-prompt graphs, calibration reduces episode-absent false positives from 58\% to 29\%.

\section{Additional evaluation metrics and controls}
\label{sec:s-extra-eval}

Object--room recall measures whether a type is filed under its correct room. Order accuracy compares the relative first-sighting order of pairs of types. Route edit distance compares remembered and true room sequences; count error compares inferred counts with distinct simulator object identities. These episode-level measures complement the frame and question scores in the main paper.

\begin{table}[t]
\centering
\caption{\textbf{Additional episode metrics.} Object--room recall and order accuracy are percentages; route edit distance is lower-is-better.}
\label{tab:s-episode}
\vspace{-0.6em}
\footnotesize
\setlength{\tabcolsep}{3.4pt}
\begin{tabular}{@{}lccc@{}}
\toprule
Graphs & Obj.--room R & Order & Route edit\\
\midrule
Baseline & \cellcolor{hourBlue!14.69}26.7 & \cellcolor{hourBlue!39.22}71.3 & 11.83\\
\quad + vocabulary map & \cellcolor{hourBlue!16.34}29.7 & \cellcolor{hourBlue!39.55}71.9 & 11.83\\
\fixes{1+2} & \cellcolor{hourBlue!29.04}52.8 & \cellcolor{hourBlue!41.09}74.7 & 10.80\\
\quad + \fix{3}: calibration & \cellcolor{hourBlue!24.91}45.3 & \cellcolor{hourBlue!43.12}78.4 & 10.80\\
\quad + door segments (ours) & \cellcolor{hourBlue!25.02}45.5 & \cellcolor{hourBlue!43.12}78.4 & \phantom{0}1.30\\
Ours, conservative calibration & \cellcolor{hourBlue!29.20}53.1 & \cellcolor{hourBlue!41.91}76.2 & \phantom{0}1.30\\
\midrule
Perfect graph & \cellcolor{hourBlue!55.00}100 & \cellcolor{hourBlue!55.00}100 & \phantom{0}0.00\\
\bottomrule
\end{tabular}
\vspace{-0.8em}
\end{table}

\begin{table}[t]
\centering
\caption{\textbf{Additional paired repair increments.} Each change compares with the preceding configuration in the main repair ladder. Fixes 1+2 compare with the vocabulary-mapped baseline. The main paper gives the door increment.}
\label{tab:s-stage}
\vspace{-0.6em}
\footnotesize
\setlength{\tabcolsep}{3.4pt}
\begin{tabular}{@{}lc@{}}
\toprule
Graphs & Step\\
\midrule
\quad + vocabulary map & \cellcolor{hourBlue!6.42}\textbf{+1.75}\ci{+0.25}{+3.25}\\
\fixes{1+2} & \cellcolor{hourBlue!26.58}\textbf{+7.25}\ci{+3.00}{+11.50}\\
\quad + \fix{3}: calibration & \cellcolor{hourPeach!3.67}$-$1.00\ci{$-$3.12}{+1.12}\\
Ours, conservative calibration & \cellcolor{hourBlue!9.17}\textbf{+2.50}\ci{+1.00}{+4.00}\\
\bottomrule
\end{tabular}
\vspace{-0.8em}
\end{table}

The intermediate per-frame-room configurations count rooms correctly in 64/200 episodes. Even the simulator graph, when rooms share type labels, counts only 138/200 exactly: perfect perception does not supply a unique physical-room identity. The repaired graph's remaining distinct-object count error is 1.08 and relation recall is 4.4\%.

\begin{table}[t]
\centering
\caption{\textbf{Content and structure controls.} Baseline-trained QA; columns give correct answers per seed, mean accuracy, and the paired change with 95\% CI.}
\label{tab:s-perturb}
\footnotesize
\setlength{\tabcolsep}{2.6pt}
\begin{tabular}{@{}lcccc@{}}
\toprule
Memory alteration & Seed 0 & Seed 1 & Acc. & $\Delta$\\
\midrule
\quad lines shuffled & 550 & 552 & \cellcolor{hourBlue!37.90}68.9 & \cellcolor{hourPeach!9.39}$\mathbf{-2.56}$\ci{$-$4.56}{$-$0.69}\\
\quad frame times removed & 530 & 534 & \cellcolor{hourBlue!36.58}66.5 & \cellcolor{hourPeach!18.11}$\mathbf{-4.94}$\ci{$-$7.44}{$-$2.44}\\
\quad other episode's memory & 501 & 512 & \cellcolor{hourBlue!34.81}63.3 & \cellcolor{hourPeach!29.77}$\mathbf{-8.12}$\ci{$-$11.12}{$-$5.12}\\
\bottomrule
\end{tabular}
\end{table}

Another episode's memory differs from the no-memory QA model by $-$1.50\ci{$-$4.00}{+1.00}, an interval including zero. This control probes useful episodic content rather than merely the presence of a memory field.

\begin{figure}[t]
\centering
\includegraphics[width=\linewidth]{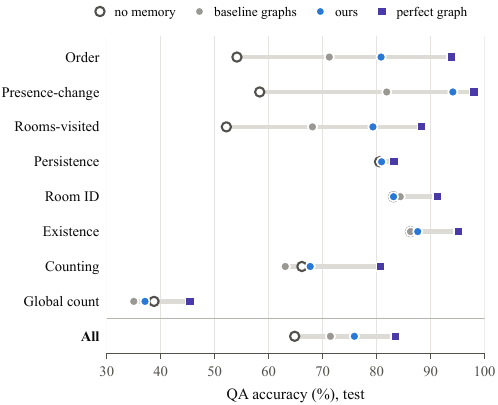}
\vspace{-1.6em}
\caption{\textbf{QA accuracy by question type after retraining} (test; two-seed means). Repaired memory helps where answerability rose (Order, Presence-change and Rooms-visited) and leaves set questions roughly unchanged; the perfect graph shows the remaining headroom on every type except Persistence.}
\label{fig:downstream}
\vspace{-0.8em}
\end{figure}

\section{Extended related-work distinctions}
\label{sec:s-related}

\para{Task and observation scope.}
Early EQA~\cite{das2018eqa,gordon2018iqa,wijmans2019eqa} couples answering to exploration, whereas OpenEQA~\cite{majumdar2024openeqa} and episodic-memory benchmarks~\cite{yadav2026findingdory,zhai2025memoryeqa,hu2025mem} emphasise retained observations. Ego4D~\cite{grauman2022ego4d} supplies egocentric video rather than a simulator's object identities; ReMEmbR~\cite{anwar2025remembr} places long-horizon memory in robot navigation. ScanQA and SQA3D~\cite{azuma2022scanqa,ma2022sqa3d} instead query reconstructed spatial scenes. These observation regimes make different oracle constructions possible. Simulator identities permit our counting audit, while the real-video experiment can test utility without the same identity-based diagnosis.

\para{What the representation preserves.}
Hierarchical and dynamic scene graphs~\cite{armeni2019sg,rosinol2020dsg,hughes2022hydra} distinguish structural levels such as rooms, places and objects; scan-level graph prediction supplies object relations~\cite{wald2020learning}. ConceptGraphs, HOV-SG and Open3DSG~\cite{gu2024conceptgraphs,werby2024hovsg,koch2024open3dsg} attach open-vocabulary semantics to spatial representations, and Clio~\cite{maggio2024clio} emphasises task-driven abstraction. Our frame-derived type lists do not inherit all of those geometric or identity properties. SayPlan, SG-Nav, GraphEQA and GraphPad~\cite{rana2023sayplan,yin2024sgnav,saxena2025grapheqa,ali2025graphpad} illustrate how a downstream reader or planner can exploit a graph. Image-memory approaches~\cite{yang2025mem,ren2024explore} retain another form of evidence; the present audit does not compare their full systems.

\para{Metric and intervention scope.}
Message-passing, motif-based, unbiased and panoptic scene-graph methods~\cite{xu2017imp,zellers2018motifs,tang2020unbiased,yang2022psg} recover object/relation annotations such as Visual Genome~\cite{krishna2017vg}. The survey of 3D scene graphs~\cite{rotondi2026survey} motivates examining downstream use beyond annotation matching. Question-prior studies~\cite{thomason2019shifting,goyal2017vqa,agrawal2018vqacp,brown2025benchmark} explain why this requires explicit controls. The relation to unanswerable reading comprehension~\cite{rajpurkar2018know} concerns missing evidence under a specified question/reader pairing, rather than a guarantee that another reader cannot answer. New Evidence, Same Choice~\cite{saha2026newevidence} also scores an intermediate evidence decision apart from the final answer: a vision-language model sees a physical measurement and must either answer or pick the cheapest experiment that resolves the question, and endpoint accuracy hides failures to switch that choice. The two diagnostics sit at different points. Theirs asks whether a model selects the evidence a decision needs; answerability asks whether memory still holds sufficient evidence before any reader acts. A memory can retain the evidence while a reader fails to select it, which is one reading of our real-video result.

The repair choices connect to separate research threads: hallucination diagnostics~\cite{rohrbach2018hallucination,li2023pope}, high-resolution visual search~\cite{wu2024vstar}, type-decision calibration rather than calibrated probabilities~\cite{guo2017calibration}, geometric room segmentation~\cite{bormann2016room}, and open-set detectors~\cite{liu2024groundingdino,minderer2023owlv2}. The controlled audit keeps the extractor fixed rather than attributing these threads' full capabilities to the repair pipeline.

\end{document}